\documentclass[runningheads]{llncs}
\usepackage{graphicx}
\usepackage{tikz}
\usepackage{comment}
\usepackage{amsmath,amssymb} 
\usepackage{eufrak,eucal}
\usepackage{color}
\usepackage{multirow}
\usepackage[normalem]{ulem}
\usepackage{booktabs}
\usepackage{hyperref}
\usepackage{tabularx}
\usepackage{wrapfig}
\useunder{\uline}{\ul}{}

\usepackage{graphicx}
\usepackage{amsmath}
\usepackage{amssymb}
\usepackage{booktabs}
\usepackage{multirow}
\usepackage{makecell}
\usepackage{tabularx}
\usepackage{multirow}
\usepackage{amssymb}
\usepackage{amsmath}
\usepackage{tabularx}
\usepackage{bm}
\usepackage[ruled,linesnumbered]{algorithm2e}
\usepackage{algorithmic}
\usepackage{xcolor}
\newcolumntype{Y}{>{\centering\arraybackslash}X}
\usepackage[capitalize]{cleveref}

\begin{document}
\pagestyle{headings}
\mainmatter
\def\ECCVSubNumber{641}  

\newcommand{\etc}{\textit{etc}}
\newcommand{\etal}{\textit{et al}}
\newcommand{\ie}{\textit{i.e.}}
\newcommand{\eg}{\textit{e.g.}}
\newcommand{\wrt}{\textit{w.r.t.} }

\title{WaveGAN: Frequency-aware GAN for High-Fidelity Few-shot Image Generation} 



\titlerunning{WaveGAN}
\author{Mengping Yang\inst{1,2} \and
Zhe Wang\inst{1,2,}\thanks{Corresponding author} \and
Ziqiu Chi\inst{1,2} \and
Wenyi Feng\inst{1,2}}

\authorrunning{M. Yang et al.}

\institute{Department of Computer Science and Engineering, East China University of Science and Technology, China \and
Key Laboratory of Smart Manufacturing in Energy Chemical Process, East China University of Science and Technology, China \\
\email{mengpingyang@mail.ecust.edu.cn} \quad
\email{wangzhe@ecust.edu.cn} \\
\email{\{chiziqiu Y10200096\}@mail.ecust.edu.cn} \quad
}

\maketitle


\begin{abstract}
\label{sec:abstract}

Existing few-shot image generation approaches typically employ fusion-based strategies, either on the image or the feature level, to produce new images.
However, previous approaches struggle to synthesize high-frequency signals with fine details, deteriorating the synthesis quality.
To address this, we propose WaveGAN, a frequency-aware model for few-shot image generation.
Concretely, we disentangle encoded features into multiple frequency components and perform low-frequency skip connections to preserve outline and structural information.
Then we alleviate the generator's struggles of synthesizing fine details by employing high-frequency skip connections, thus providing informative frequency information to the generator.
Moreover, we utilize a frequency $L_1$-loss on the generated and real images to further impede frequency information loss.
Extensive experiments demonstrate the effectiveness and advancement of our method on three datasets.
Noticeably, we achieve new state-of-the-art with FID 42.17, LPIPS 0.3868, FID 30.35, LPIPS 0.5076, and FID 4.96, LPIPS 0.3822 respectively on Flower, Animal Faces, and VGGFace.
GitHub: \textcolor{red}{\url{https://github.com/kobeshegu/ECCV2022_WaveGAN}}



\end{abstract}


\section{Introduction}
\label{sec:introduction}
Recent years have witnessed remarkable developments in visual generative tasks with the rapid progress of generative models, especially Generative Adversarial Networks (GANs)~\cite{karras2018progressive}~\cite{karras2019style}~\cite{karras2020analyzing}~\cite{Karras2021}.
Despite being applied to various domains, the success of GANs mainly comes from immense data, and GANs struggle to synthesize high-quality images given insufficient data.
Few-shot learning~\cite{fei2006one}, being proposed to improve the generalization ability in the limited-data scenarios, has gained extensive attention and focused research.
However, most of existing few-shot algorithms are designed for classification~\cite{liu2020negative} and segmentation~\cite{li2021adaptive} problems, few studies address few-shot image generation.
Therefore, exploring and facilitating the generation quality in the few-shot regime is necessary.

Few-shot image generation aims at generating novel images for a category when given a few images from the same category.
Specifically, the model is first trained on an auxiliary dataset (seen classes) with sufficient data in an episode-based manner~\cite{vinyals2016matching}, \emph{i.e.}, feeding a specific number of images (\eg, 3, 5) into the model in each episode.
The trained model is then expected to produce new images when given a few images of a category from a new dataset (unseen classes).
There is no overlap between the auxiliary dataset and the new testing dataset.
The model is encouraged to capture the transferable ability from seen classes to unseen classes to generate new images for unseen classes.

Previous methods try to 1) transform intra-class information~\cite{antoniou2017data}, 2) design new optimization schemes by combining GANs with meta-learning~\cite{clouatre2019figr}~\cite{liang2020dawson}, and 3) fuse the given images~\cite{hong2020matchinggan}~\cite{hong2020f2gan}~\cite{gu2021lofgan} to address few-shot image generation.
Among these methods, LoFGAN~\cite{gu2021lofgan} achieves current state-of-the-art performance by fusing local representations based on the semantic similarity of features.
However, existing approaches ignore the enormous impact of the frequency information throughout the generating process.
F-principal~\cite{xu2019frequency} proves that neural networks preferentially fit frequency signals from low to high.
Consequently, the model tends to generate frequencies with higher priority and more superficial complexity, \emph{i.e.}, only generating low-frequency signals.
\begin{figure}
	\centering
	\includegraphics[width=\linewidth]{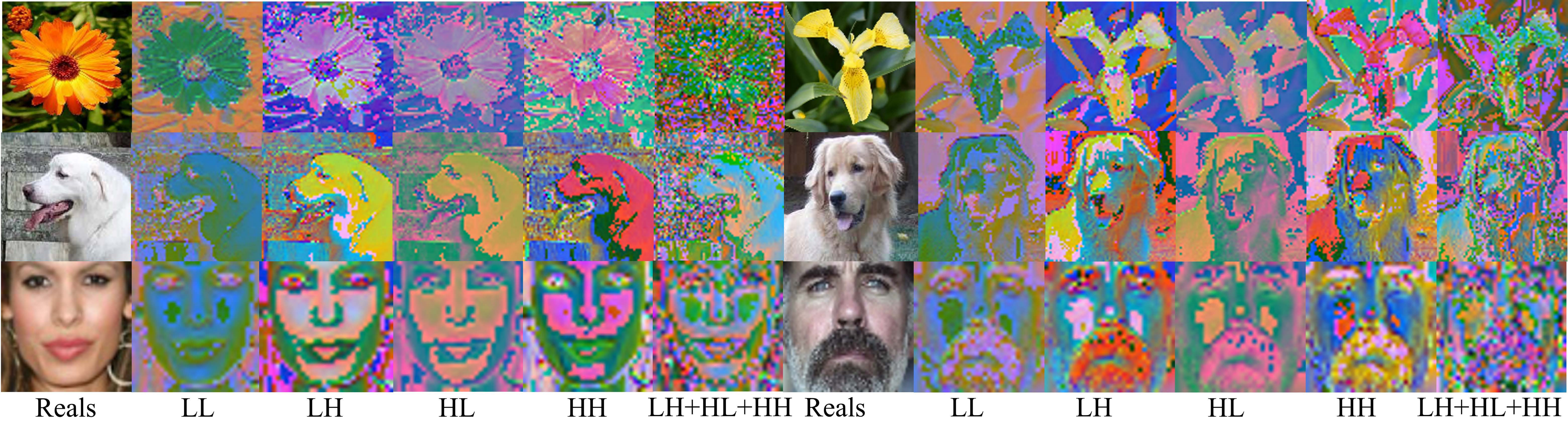}
	\caption{\textbf{Visualization of the transformed frequency components.} $LL$ denotes the low-frequency component, and $LH, HL, HH$ represent the high-frequency components.}
	\label{fig:WaveletTrans}
\end{figure}

We visualize different frequency components of real images in Fig~\ref{fig:WaveletTrans}.
The low-frequency component (\emph{i.e.}, $LL$) contains general information like the overall surface, outline, and structure.
While rich details and perceptible information like the leaves of flowers, the tongue of dogs, and the hair of human, lie in the high-frequency components (\emph{i.e.}, $LH$, $HL$, $HH$).
Rich details can be obtained by adding all high-frequency components (\emph{i.e.}, $LH + HL + HH$) together.
Since high frequency components contain meticulous information, loss of them may lead the generator to synthesize blurry images with more aliasing artifacts.
This issue highlights the necessity of considering frequency signals in generating images, especially high-frequency signals as the generator usually eschews them~\cite{jiang2021focal}~\cite{xu2019frequency}.

This paper proposes WaveGAN, an innovative and effective approach to ameliorate the few-shot synthesis quality from the perspective of frequency domains.
We first perform wavelet decomposition to transform the encoded features from the spatial domain to multiple frequency domains, comprising low and high-frequency components.
Then we feed the low-frequency component to the behind layers of the encoder via low-frequency skip connections, maintaining the overall outline and structural patterns.
To mitigate the pressure of the generator to generate high-frequency signals and provide more details to the decoder, we directly feed the decomposed high-frequency signals to the decoder.
Two strategies are designed to aggregate the high-frequency signals, namely WaveGAN-M and WaveGAN-B.
Both of them are effective and can provide high-frequency information to the decoder.
The high-frequency components are then precisely reconstructed back into the original features with our inverse frequency transformation operations, guaranteeing the minimal loss of high-frequency signals.
In addition, we apply frequency $L_1$-loss to the generated images and the real images, which is complementary to spatial losses and impedes losing frequency information.
Our primary contributions can be summarized as follows:
\begin{itemize}
  \item We propose WaveGAN, the first few-shot image generation method that exploits frequency components to promote synthesis quality.  
  Adding low and high-frequency skip connections to the generator, our WaveGAN alleviates the generator's struggles to encode high-frequency signals and provide more perceptible information, resulting in favorable generation quality.
  \item We design two techniques to aggregate the high-frequency information for reconstructing frequency signals back to the original features, \emph{i.e.}, WaveGAN-M and WaveGAN-B, which preserve fine details and statistical properties. We also present frequency $L_1$-loss to avoid losing frequency information.
  \item We conduct comprehensive experiments on three datasets. Both qualitative and quantitative results demonstrate the superiority and effectiveness of our method.
  Notably, our model outperforms the state-of-the-art approach with significant FID improvements (\emph{e.g.}, from \textbf{102.07} to \textbf{30.35} on Animal Face).
\end{itemize}


\section{Related Work}
\label{sec:relatedwork}

\textbf{Generative Adversarial Network.}
Generative Adversarial Networks (GANs) have made significant progress since the pioneering work in~\cite{goodfellow2014generative}.
Benefit from remarkable ability of capturing the data distribution, GANs have been successfully applied in various visual domains, including image generation~\cite{karras2020analyzing}~\cite{Karras2021}, video generation~\cite{wang2019event}, image-to-image translation~\cite{park2020contrastive}~\cite{richardson2021encoding}, etc.
Typically, a GAN model consists of a generator and a discriminator, and the two networks are updated alternatively in an adversarial manner.
Training a GAN is notoriously formidable as it requires massive data and computation resources, and the adversarial training may make the model diverge.
The discriminator is tend to overfitting when given limited data, resulting in poor generation quality.
Several works have been proposed to mitigate the discriminator overfitting.
Different data augmentation techniques, including differentiable~\cite{DiffAug}, non-leaking~\cite{karras2020training} and adaptive pseudo augmentation~\cite{jiang2021deceive} are designed to expand the limited training data.
Lecam~\cite{tseng2021regularizing} regularizes the output of the discriminator to avoid overfitting.
Unlike these efforts made for unconditional image generation with limited data, in this paper, we seek to generate novel images for one specific category when given a few images from this category.

\textbf{Wavelet Transformation in GANs.}
Decomposing given signals into different frequency components, wavelet transformation has made great success in various generative tasks such as style transfer~\cite{yoo2019photorealistic}, image reconstruction~\cite{jiang2021focal}, image inpainting~\cite{yu2021wavefill}, image editing~\cite{gao2021high} and image super-resolution~\cite{deng2019wavelet}~\cite{huang2019wavelet}.
These approaches try to narrow the information gaps in the frequency domain to boost the model's performance.
For example, Jiang \emph{et al.} propose focal frequency loss to avoid the loss of important frequency information for image reconstruction tasks~\cite{jiang2021focal}.
WaveFill~\cite{yu2021wavefill} decomposes images into multiple frequency components and fills the corrupted image regions with decomposed signals, which achieves superior image inpainting.
Different from these methods, we try to generate realistic and plausible images when given only a few data.
We are interested in the influence of frequency information on the challenging few-shot image generation.

\textbf{Few-shot Image Generation.}
Inspired by the human's great generalization ability from a few observations, few-shot image generation models try to generate new images given a few images.
Existing few-shot image generation approaches can be roughly divided into three categories: 1) Optimization-based, 2) Fusion-based, and 3) Transformation-base methods.
DAGAN~\cite{antoniou2017data} transforms combined projected latent codes and encoded images to new images.
The optimized-based methods FIGR~\cite{clouatre2019figr} and DAWSON~\cite{liang2020dawson} combine generative models with optimization-based meta learning Reptile~\cite{nichol2018reptile} and MAML~\cite{finn2017model}, respectively.
The fusion-based methods fuse the local feature~\cite{gu2021lofgan} or the input images~\cite{hong2020f2gan} ~\cite{hong2020matchinggan} to synthesis novel images.
GMN~\cite{bartunov2018few} combines VAE~\cite{kingma2013auto} with Matching Networks~\cite{vinyals2016matching} to capture the few-shot distribution.
MatchingGAN~\cite{hong2020matchinggan} matches random vectors with given real images and mapping the fused features to novel images.
F2GAN~\cite{hong2020f2gan} further improves MatchingGAN with a fusing-and-filling paradigm.
By fusing local representations with semantic similarity, LoFGAN~\cite{gu2021lofgan} promotes the generation quality.
Notably, zero-shot or few-shot text-to-image generation methods~\cite{ramesh2021zero}~\cite{gu2022vector}~\cite{ramesh2022hierarchical} have made great progress recently. Differently, this paper focuses on the problem of few-shot image generation for generating new images for a given class as defined in Sec.~\ref{sec:methodoverview}.

However, existing methods ignore the influence of frequency components on the quality of generated images, leading the generator to synthesize unfavorable images with more artifacts and fewer details.
In this paper, we present a frequency-aware model that can generate appealing and photorealistic images by adding low and high-frequency skip connections to the generator.
Such design mitigates the generator's pressure of synthesising high-frequency signals.
Our work explores an effective solution for few-shot image generation from the frequency domain perspective, which complements previous fusion-based methods.




\section{Methodology}
\label{sec:method}

\subsection{Overview}
\label{sec:methodoverview}
\textbf{Problem Definition.}
Given $K$ images from a new class, our model's goal is to synthesis diverse and plausible images for the given class.
The number of images $K$ defines a $K$-shot image generation task.
Generally, this task is accomplished in two phases, \emph{i.e.}, training and testing.
The datasets is first split into seen classes $\mathbb{C}_s$ and unseen classes $\mathbb{C}_u$, where $\mathbb{C}_s$ and $\mathbb{C}_u$ have no overlap.
In the training phase, a substantial amount of $K$-shot image generation tasks sampled from $\mathbb{C}_s$ are fed into the model, expecting the model to transfer the knowledge of generating new images learned from $\mathbb{C}_s$ to $\mathbb{C}_u$.
In the testing phase, the model takes images from $\mathbb{C}_u$ as input to synthesis new images.
\begin{figure}
	\centering
	\includegraphics[width=\linewidth]{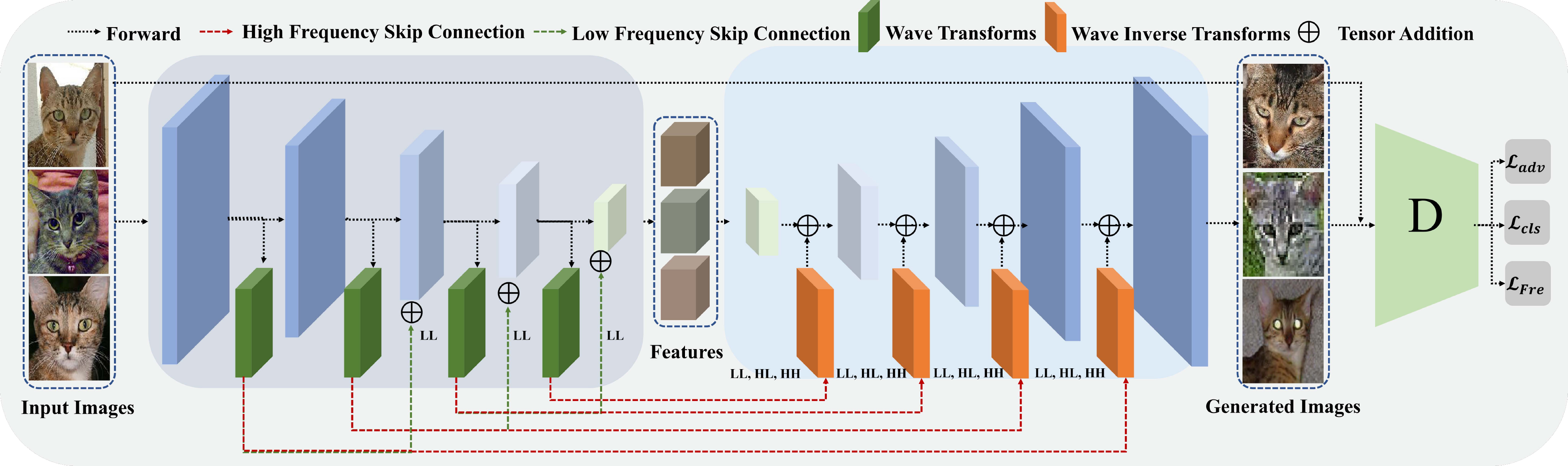}
	\caption{\textbf{The overall framework of our WaveGAN.} We employ low frequency ($LL$) skip connections in the encoder and high frequency skip connection ($LH, HL, HH$) in the decoder to provide rich details to improve synthesis quality.}
	\label{fig:framework}
\end{figure}

\textbf{Overall Framework.}
As shown in Fig.~\ref{fig:framework}, our model consists of a WaveEncoder, a WaveDecoder, and a Discriminator, the WaveEncoder and Wavedecoder constitute our generator.
The WaveEncoder extracts feature representations of images, while the WaveDecoder maps the feature representation to new images.
We perform wavelet transformation to the encoded features and obtain multiple frequency components.
Then we employ low-frequency skip connections in the encoder to preserve the overall structure and outline.
We exploit high-frequency skip connections to provide detailed information to the decoder.
The wavelet inverse transformation module reconstructs these high-frequency signals to the original features.
The high-frequency signals contain rich details and perceptible information, enabling the generator to synthesis high-quality images.
The real and generated images are then fed into the discriminator to train the whole model.
Next, we elaborate our WaveEncoder and WaveDecoder in detail.


\subsection{WaveEncoder}
Our WaveEncoder is composed of convolutional blocks and wavelet transformation blocks.
The convolutional operations extract features for the decoder to produce new images.
To disentangle the extracted features into multiple frequency components, we adopt a simple yet effective wavelet transformation, \emph{i.e.}, Haar wavelet~\cite{daubechies1990wavelet}.
Haar wavelet contains two operations: wavelet transform and inverse wavelet transformation, and four kernels, namely $LL^T$, $LH^T$, $HL^T$, and $HH^T$.
\begin{equation}
L^{\top}=\frac{1}{\sqrt{2}}\left[\begin{array}{ll}
1 & 1
\end{array}\right], \quad H^{\top}=\frac{1}{\sqrt{2}}\left[\begin{array}{ll}
-1 & 1
\end{array}\right]
\end{equation}
where $L$ and $H$ denote the low and high pass filters, respectively.
The low pass filter focuses on low-frequency signals containing the outline and structural information.
In contrast, the high pass filter emphasizes high-frequency signals that capture fine-grained details like subtle edges and contour (See Fig.~\ref{fig:WaveletTrans} and~\ref{fig:Introduction}).

The wavelet transformation decomposes features into frequency components $LL$, $LH$, $HL$, $HH$.
Among these frequency signals, $LL$ captures the overall appearance and basic object structures of images (See Fig.~\ref{fig:WaveletTrans}).
Thus we employ low-frequency skip connections in the encoder to obtain precise and faithful features throughout the feature extracting process.
Specifically, for feature $E_i$ obtained from the $i$-th convolutional block in the encoder, we adopt Haar wavelet transformation to extract the frequency components $LL_i$, $LH_i$, $HL_i$, $HH_i$.
We then perform tenor addition on the low frequency signal $LL_i$ and feature $E_{i+1}$ obtained from the $(i+1)$-th convolutional block in the encoder.
\begin{equation}
E_{i+2} = LL_i + ConvBlock_{i+1}(E_{i+1})
\end{equation}
The obtained skip connected feature $E_{i+2}$ is fed into the $(i+2)$-th convolutional block. The low-frequency skip connections contribute to the fidelity of the generated images.
The experimental proofs are given in Sec.~\ref{sec:ablation}.

\subsection{WaveDecoder}
High-frequency components contain rich details of images.
However, deep networks usually fit frequency signals from low to high, making it difficult for the generator to produce high-frequency information since it generates frequencies with higher priority.
To alleviate the encoder's pressure to synthesis rich details and provide fine-grained information to the decoder, we directly feed the decomposed high-frequency signals $LH$, $HL$, and $HH$ into the WaveDecoder via high-frequency skip connections.
Specifically, for the $i$-th layer of the encoder, we perform wavelet transformation on the features and obtain high frequency components $LH_i$, $HL_i$, $HH_i$, then we feed the inversed components to the $(n-i)$-th layer of the decoder as exhibited in Fig.~\ref{fig:framework}, where $n$ is the number of all layers.
This operation encourages the decoder to synthesis images with more details and fewer artifacts.
We employ wavelet inverse transformation to reconstruct high-frequency signals back to original features.
Our wavelet inverse transformation can be categorized into Mean and Base-index inverse transformation based on how these high-frequency components are aggregated.

\begin{figure}
	\centering
	\includegraphics[width=.6\linewidth]{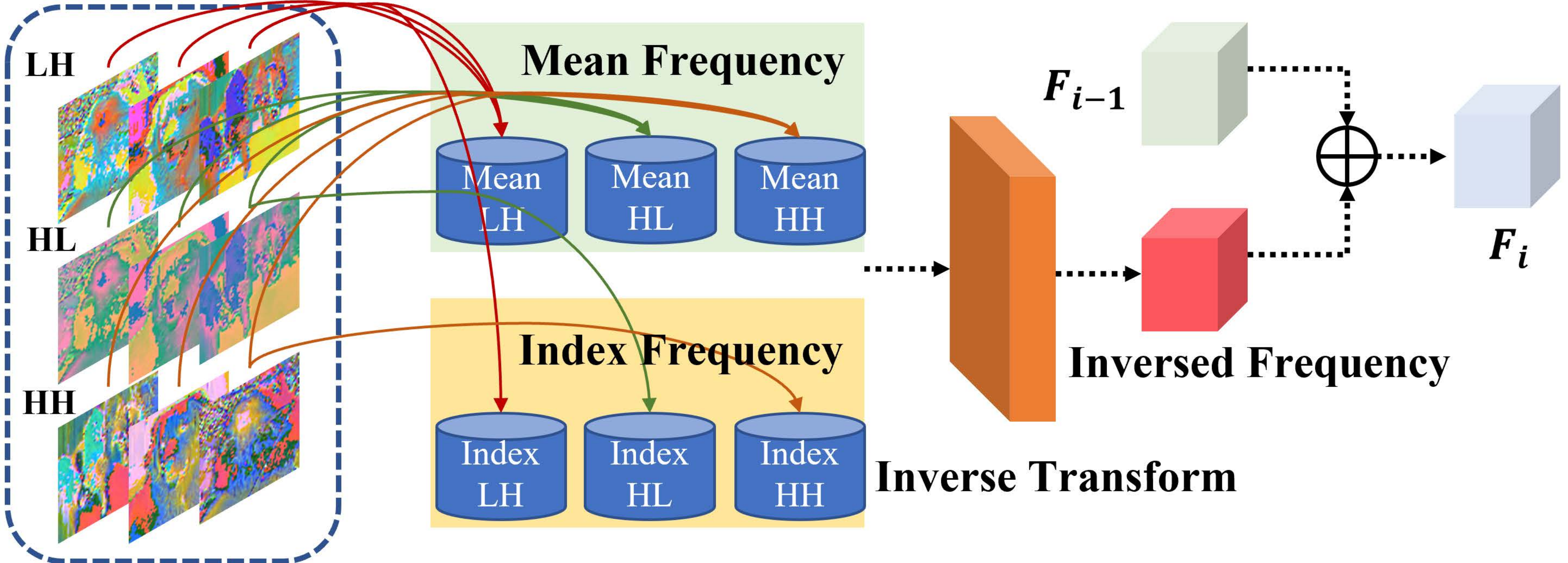}
	\caption{\textbf{Illustration of our Mean and Base-index inverse transformation.}}
	\label{fig:MorBfrequency}
\end{figure}

\textbf{Mean Inverse Transformation.}
As presented in Fig.~\ref{fig:MorBfrequency}, we calculate the frequency element-wise average of all high-frequency components of $K$ features from the same category, and take the averaged results as the input of our Mean inverse transformation module.
\begin{equation}
HF_M = \sum\limits_{{{i = }}1}^{{K}} HF_i, HF_i \in \{LH_i, HL_i, HH_i\}
\end{equation}

Although providing high-frequency signals to the decoder facilitates the generated images' quality, the Mean inverse transformation may not be suited for one specific image as the averaged frequency information may shift the frequency signals.
The averaged frequency information becomes more neutral with the number of training images $K$ increases, leading to a decrease in generalization.
This conjecture may go against our common sense that the generalization ability should improve as the number of images increases.
We analyze this is because different images, even from the same category, have different frequency signals in the frequency domain.
The experiments in Sec.~\ref{sec:experiments} confirm our analysis that the averaged frequency transformation may fail to generalize when $K$ is bigger.
To improve the generalization ability of our inverse transformation in the frequency domain, we design shots-agnostic Base-index inverse transformation.

\textbf{Base-Index Inverse Transformation.}
Our Base-index inverse transformation is implemented based on the Local representation Fusion (LoF) strategy of LoFGAN~\cite{gu2021lofgan}.
We first give a brief introduction to LoFGAN.
Given encoder features $\mathbb{F} = \mathbb{E}(X) \in \mathbb{R}^{k \times w \times h \times c}$, LoFGAN randomly selects one base feature ${f}_{base} \in \mathbb{R}^{w \times h \times c}$ and views the rest $(K-1)$ features as reference, \emph{i.e.}, ${f}_{ref} \in \mathbb{R}^{(k-1) \times w \times h \times c}$.
LoFGAN fuses the local features based on the calculated semantic similarity map and replaces the closest base feature with the fused feature.

To provide customized high-frequency signals to the generated images, we recode the index of the selected base feature of the LoF module.
As illustrated in Fig.~\ref{fig:MorBfrequency}, instead of calculating the averaged frequency signals, we explicitly feed the high-frequency component corresponding to the recorded index $i$.
\begin{equation}
HF_B = HF_i, HF_i \in \{LH_i, HL_i, HH_i\}
\end{equation}
The high-frequency components are exact frequency signals of the selected feature, thus providing customized rich details and perceptible information to the decoder.
The generalization ability of our Base-index inverse transformation would not deteriorate with the training number increases.

After obtaining the aggregate high-frequency signals $LH$, $LH$, $HH$, we perform our inverse transformation to reconstruct these signals back to the original features.
Concretely, we first perform transposed-convolution on each frequency component, then sum up all the output features.
The summed result convert frequency signals back to original features precisely, theoretical analysis can be found in~\cite{yoo2019photorealistic}.
The inverse transformation can be formally expressed as:
\begin{equation}
{F_{IF}} = \sum {TransConv(HF)} ,HF \in \{ HF_M , HF_B\}
\end{equation}
$HF$ is obtained from either the Mean or the Base-index inverse transformation. Element-wise tensor addition are employed to integrate the inversed high-frequency components with the former feature, \emph{i.e.}, $F_i = F_{i-1} + F_{IF(n-i+1)}$.
Such branches assuage the generator's dilemmas in producing limited high-frequency containing fine details.

\subsection{Optimization Objective}
Our model has two networks to optimize, the generator (G) and the discriminator (D).
The input of G is real images, and G tries to generate plausible and diverse new images $\hat{x} = G(X)$.
Let $X = \sum\limits_{i = 1}^K {{x_i}}$ denotes the real images, and $c(x_i)$ denotes the labels for the image $x_i$ (available for the $\mathbb{C}_s$ only).
The inputs of D are the real and generated images, and D tries to distinguish the real images from the generated ones.
The generator and the discriminator are updated alternatively in an adversarial manner by optimizing the following losses.

\textbf{Frequency $L_1$-Loss.}
We employ the frequency $L_1$-loss on the transformed frequency components of generated images and the real images, impeding losing the frequency information.
Besides, the frequency loss complements existing spatial losses. We perform wavelet transformation on the generated and real images to compute the frequency $L_1$ loss.
\begin{equation}
{{\cal L}_{Fre}} = \sum {{{\left\| {Fre_{x} - Fre_{\hat{x}}} \right\|}_1},Fre \in } \{ LL,LH,HL,HH\}
\end{equation}

\textbf{Local Reconstruction Loss.}
We adopt local reconstruction loss to constrain the model to maintain the local features.
\begin{equation}
\mathcal{L}_{{rec }}=\|\hat{x}-\operatorname{LFM}(X, \boldsymbol{\alpha})\|_{1}
\end{equation}
where $\boldsymbol{\alpha}$ denotes the coefficient vector to fuse the features in the local fusion module (LoF).

\textbf{Adversarial Loss.}
Following~\cite{gu2021lofgan} and ~\cite{odena2017conditional}, we adopt the hinge version of adversarial loss to optimize the generator and the discriminator.
\begin{equation}
\begin{aligned}
\mathcal{L}_{{adv}}^{D} &=\max (0,1-D(x))+\max (0,1+D(\hat{x})) \\
\mathcal{L}_{{adv}}^{G} &=-D(\hat{x})
\end{aligned}
\end{equation}

\textbf{Classification Loss.}
Classification loss constrains the model to synthesis images that belong to one specific category.
We add an auxiliary classifier to the generator and the discriminator following ACGAN~\cite{odena2017conditional}.
The classification loss encourages the discriminator to identify which category an image belongs to while enabling the generator to synthesis images that belong to one specific category.
 \begin{equation}
\begin{aligned}
&\mathcal{L}_{{cls}}^{D}=-\log P(c(x) \mid x) \\
&\mathcal{L}_{{cls}}^{G}=-\log P(c(\hat{x}) \mid \hat{x})
\end{aligned}
\end{equation}

Our model is optimized with the following objective function with the linear combination of the above losses.
\begin{equation}
\begin{aligned}
{{\cal L}_G} &= {\cal L}_{{{adv }}}^G + \lambda _{{{cls }}}^G{\cal L}_{{cls }}^G + {\lambda _{Fre}}{{\cal L}_{Fre}} + {\lambda _{{rec }}}{\cal L}_{{rec }}^G  \\
\mathcal{L}_{D} &=\mathcal{L}_{{adv}}^{D}+\lambda_{{cls}}^{D} \mathcal{L}_{{cls}}^{D}
\end{aligned}
\end{equation}


\section{Experiments}
\label{sec:experiments}
\textbf{Datasets.}
We use three popular datasets in the few-shot image generation community to evaluate the performance of our model, namely Flower~\cite{nilsback2008automated}, Animal Faces~\cite{liu2019few} and VGGFace~\cite{cao2018vggface2}.
These datasets are split into seen classes $\mathbb{C}_s$ and unseen classes $\mathbb{C}_u$.
$\mathbb{C}_s$ is used in the training stage, while $\mathbb{C}_u$ is used in the testing stage.
The adopted datasets are split in Tab.~\ref{tab:datasets} following~\cite{gu2021lofgan} and~\cite{hong2020f2gan}.

\begin{table}
\centering
\caption{\textbf{The split of experimental datasets.} The seen classes are used for training and the unseen classes are used for testing. }
\resizebox{.7\textwidth}{!}{
\begin{tabular}{c|c|c|c|c}
\toprule
Datasets & \#Total classes & \#Seen classes & \#Unseen classes & \#Images/class \\ \hline
Flower   & 102           & 85           & 17             & 40           \\
Animal Faces   & 149           & 119          & 30             & 100          \\
VGGFace  & 2354          & 1802         & 552            & 100          \\ \bottomrule
\end{tabular}
}
\label{tab:datasets}
\end{table}

\textbf{Evaluation and Baselines.}
We evaluate the quality of the generated images with two commonly used metrics: Fr\'{e}chet Inception Distance (FID)~\cite{FIDnips2017} and Learned Perceptual Image Patch Similarity (LPIPS)~\cite{zhang2018unreasonable}. The two metrics are calculated in the same setting with~\cite{gu2021lofgan}.
We compare our model with several few-shot image generation approaches, namely FIGR~\cite{clouatre2019figr}, GMN~\cite{bartunov2018few}, DAWSON~\cite{liang2020dawson}, DAGAN~\cite{antoniou2017data}, MathingGAN~\cite{hong2020matchinggan}, F2GAN~\cite{hong2020f2gan} and LoFGAN~\cite{gu2021lofgan}.
We re-implement the current state-of-the-art LoFGAN for fair comparisons (denoted as ``LoFGAN{$^\ddagger$}''), and all methods are evaluated under the same conditions.

\subsection{Quantitative Evaluation}
We first train the model with $\mathbb{C}_s$ and then use the data from $\mathbb{C}_u$ to synthesis novel images for quantitative evaluation.
Following LoFGAN~\cite{gu2021lofgan} and~\cite{hong2020f2gan}, we split each unseen class into two parts, $\mathbb{S}_{sup}$ and $\mathbb{S}_{que}$, the images in $\mathbb{S}_{sup}$ are fed into the model to generate images.
We generate 128 images for each class (denoted as $\mathbb{S}_{gen}$), $\mathbb{S}_{gen}$ and $\mathbb{S}_{que}$ are used to compute FID (lower is better) and LPIPS (higher is better) scores to evaluate the synthesis quality.
The quantitative results of our model and the baselines are given in Tab.~\ref{tab:quantitativeall}.
All results in the table are conducted under 3-shot setting for both training and testing stages.

As can be observed from Tab.~\ref{tab:quantitativeall}, our waveGAN achieves the lowest FID and the highest LPIPS on all the datasets, and both WaveGAN-M and WaveGAN-B outperform the baseline models.
Notably, Our model achieves much better FID results than the current state-of-the-art LoFGAN.
Specifically, WaveGAN-B achieves the FID of less than 5 (\textbf{4.96}) on the challenging VGGFace dataset, while LoFGAN obtains 16.82. And WaveGAN-B lowers the FID from 102.07 (\emph{resp.}, 81.70) to \textbf{30.35} (\emph{resp.}, \textbf{42.17}) on Animal Faces (\emph{resp.}, Flower).
Such significant improvements on the quantitative metrics demonstrate that our model could generate plausible and vivid images.
Since the upper bound of FID scores equals 0 and we achieve a single digit for the first time, demonstrating the efficacy of our method.
As for the LPIPS metric, we calculate its upper bound by measuring the real images' LPIPS score and obtain 0.4393 for Flower, 0.5729 for Animal Faces, and 0.4389 for VGGFace.
Our model yields favorable LPIPS scores that approach the upper bound, further substantiating the effectiveness of our model.
\begin{table}[]
\centering
\caption{\textbf{Quantitative comparison results of our model and the baselines on FID and LPIPS}. {$^\dagger$} results are quoted from LoFGAN~\cite{gu2021lofgan}. {$^\ddagger$} results are re-impelemented under the same condition with our model for fair comparison. The best and the second-ranked results are \textbf{bold} and {\ul underlined}, respectively.}
\resizebox{\textwidth}{!}{
\begin{tabular}{cccccccc}
\toprule
\multirow{2}{*}{Method} & \multirow{2}{*}{Type} & \multicolumn{2}{c}{Flowers}      & \multicolumn{2}{c}{Animal Faces} & \multicolumn{2}{c}{VGGFace}     \\
                        &                       & FID ({$\downarrow$}) & LPIPS ({$\uparrow$}) & FID ({$\downarrow$}) & LPIPS ({$\uparrow$})   & FID ({$\downarrow$})  & LPIPS ({$\uparrow$})        \\ \midrule
FIGR{$^\dagger$}~\cite{clouatre2019figr}                    & Optimization          & 190.12         & 0.0634          & 211.54         & 0.0756          & 139.83        & 0.0834          \\
DAWSON{$^\dagger$}~\cite{liang2020dawson}                   & Optimization          & 188.96         & 0.0583          & 208.68         & 0.0642          & 137.82        & 0.0769          \\
DAGAN{$^\dagger$}~\cite{antoniou2017data}                   & Transformation        & 151.21         & 0.0812          & 155.29         & 0.0892          & 128.34        & 0.0913          \\
GMN{$^\dagger$}~\cite{bartunov2018few}                      & Fusion                & 200.11         & 0.0743          & 220.45         & 0.0868          & 136.21        & 0.0902          \\
MatchingGAN{$^\dagger$}~\cite{hong2020matchinggan}          & Fusion                & 143.35         & 0.1627          & 148.52         & 0.1514          & 118.62        & 0.1695          \\
F2GAN{$^\dagger$}~\cite{hong2020f2gan}                      & Fusion                & 120.48         & 0.2172          & 117.74         & 0.1831          & 109.16        & 0.2125          \\
MatchingGAN+LoFGAN{$^\dagger$}~\cite{gu2021lofgan}          & Fusion                & 86.59          & 0.3704          & 112.99         & {\ul 0.5024}    & 22.99         & 0.2687          \\
LoFGAN{$^\dagger$}~\cite{gu2021lofgan}                      & Fusion                & 79.33          & {\ul0.3862}     & 112.81         & 0.4964          & 20.31         & 0.2869          \\ \midrule
LoFGAN{$^\ddagger$}                                         & Fusion                & 81.70          & 0.3768          & 102.07         & 0.5005          & 16.82         & 0.3041          \\
\textbf{WaveGAN-M (Ours)}                                   & Fusion                & {\ul 63.79}    & 0.3709          & {\ul 50.98}    & 0.5014          & {\ul 8.62}    & \textbf{0.3822} \\
\textbf{WaveGAN-B (Ours)}                                   & Fusion                & \textbf{42.17} & \textbf {0.3868}& \textbf{30.35} & \textbf{0.5076} & \textbf{4.96} & {\ul 0.3255}    \\ \bottomrule
\end{tabular}
}
\label{tab:quantitativeall}
\end{table}

\subsection{Qualitative Evaluation}
\begin{figure}
	\centering
	\includegraphics[width=\linewidth]{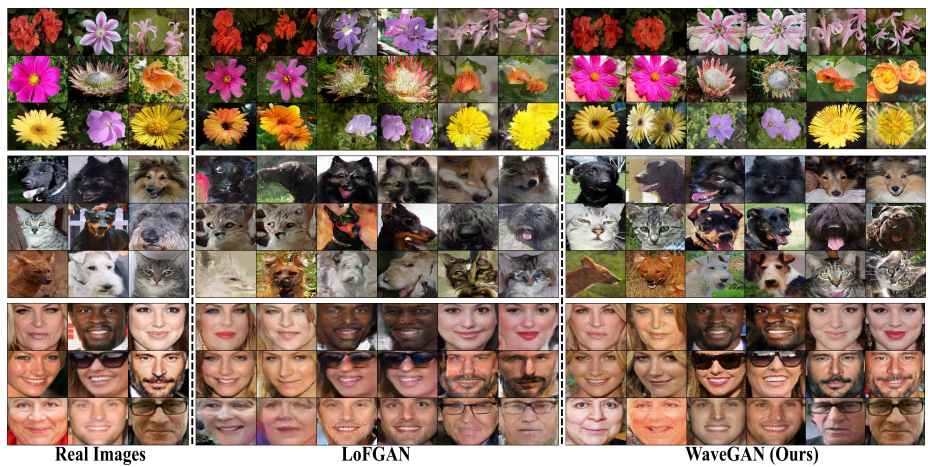}
	\caption{\textbf{Qualitative comparison results of our WaveGAN and LoFGAN.} The left-most three columns are real images, we give two generated images for each class of real image.}
	\label{fig:qualitative}
\end{figure}

We present the visualization results of LoFGAN~\cite{gu2021lofgan} and our WaveGAN-B for qualitative comparison in Fig.~\ref{fig:qualitative}. For each real image, we give two fake images generated by LoFGAN and our WaveGAN-B.
As can be observed from the figure, images generated by our model are more plausible than that of LoFGAN.
Moreover, images generated by our model contain rich details and perceptible information.
Take the generated flowers as examples, the horizontal and vertical orientation of petals, the details of stamens, and the shapes of the leaves of images synthesised by our model are more reasonable and realistic.
Moreover, animal and human face images generated by LoFGAN are distorted with blurry unfavorable artifacts, features like the eyes of cats, hair of dogs are even misplaced.
By contrast, animal and human face images generated by our model have higher fidelity and even look indistinguishable from real images.

\subsection{Visualization of the frequency components of generated images}
\begin{figure}
	\centering
	\includegraphics[width=.8\linewidth]{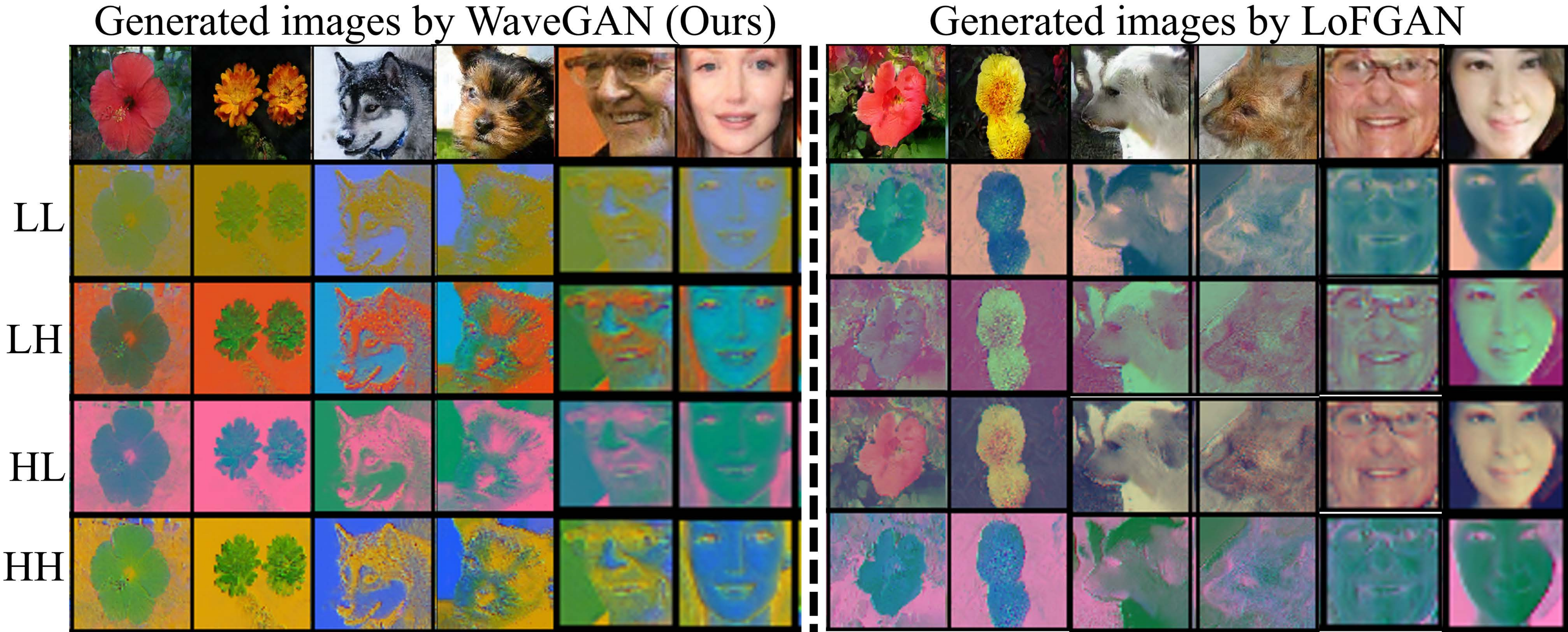}
	\caption{\textbf{Visualization results of the frequency components of images generated by our WaveGAN and LoFGAN}~\cite{gu2021lofgan}.}
	\label{fig:Introduction}
\end{figure}

We visualize the frequency components of images generated by our WaveGAN and LoFGAN~\cite{gu2021lofgan} in Fig.~\ref{fig:Introduction}.
As observable in Fig.~\ref{fig:Introduction}, in addition to the fact that our WaveGAN produces more realistic visual images, the decomposed high-frequency components of our WaveGAN contain more details and perceptible information than LoFGAN.
Specifically, the frequency components of LoFGAN contain only the surface and texture information of images, indicating that the generator of LoFGAN fails to synthesize high-frequency information.
In comparison, our WaveGAN is frequency-aware and can produce high-frequency signals that contain more fine details and statistical properties.
Further, the frequency components of WaveGAN capture delicate information that is less noticeable (\emph{e.g.}, the glasses frames and flower rhizomes in the second and fifth row of Fig.~\ref{fig:Introduction}, respectively).
Such observation further demonstrates the effectiveness and advancement of our method.

\subsection{Ablation Studies}
\label{sec:ablation}
We conduct ablation studies to evaluate the effectiveness of each component of the proposed WaveGAN.
There are three main components of our WaveGAN, namely 1) the low-frequency skip connection, 2) the high-frequency skip connection, and 3) frequency $L_1$-loss.
We remove each component and keep other settings unchanged to validate their contributions.
Besides, we remove the LoF module to investigate the influence of local fusion on our model.
We also test the contribution of each component for our two transformation techniques (\emph{i.e.}, WaveGAN-M and WaveGAN-B) in the appendix.

\begin{figure}
	\centering
	\includegraphics[width=.8\linewidth]{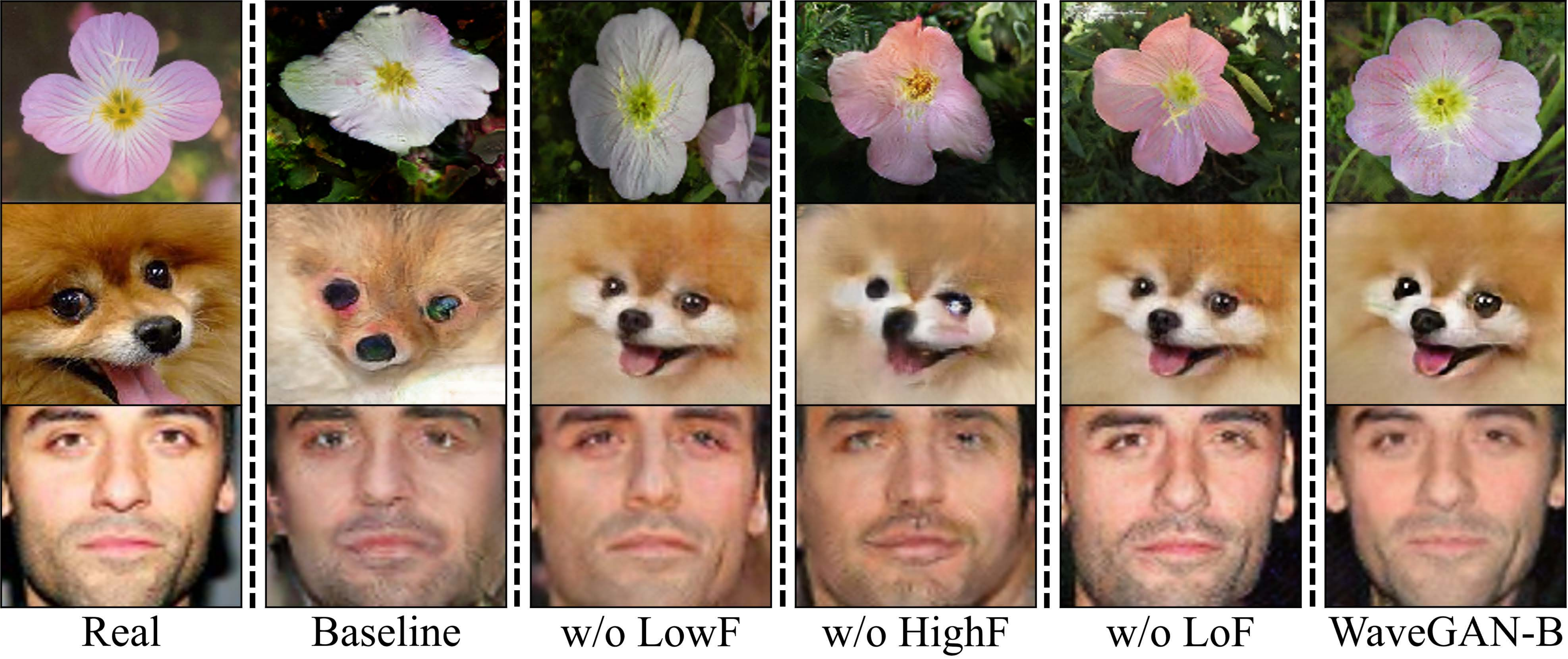}
	\caption{\textbf{Visualization comparison results of ablation studies.}}
	\label{fig:ablation}
\end{figure}
We give visualization results of ablation studies in Fig.~\ref{fig:ablation}.
The quantitative results of ablation studies are given in Sec.~\ref{sec:AblationApp} of the appendix.
Combining the qualitative and quantitative results, we can draw four conclusions as follows:
1) The skip connection of both low and high-frequency signals play an essential role in our model, and our high-frequency skip connections play a predominant role in our model. 
2) High-frequency information provides detailed information to the generated images, and low-frequency information provides the overall outline of images (compare baseline with other generated images in Fig.~\ref{fig:ablation}).
3) Our full model achieves the best results and can generate satisfactory images.
4) Our method complements to the local feature fusion approach.

\subsection{Augmentation for Classification}
To further investigate the quality of generated images, we augment the datasets with images generated by our model for downstream image classification tasks.
Specifically, we first pre-train a ResNet18 network with seen classes following~\cite{gu2021lofgan} and~\cite{hong2020f2gan}, we train the ResNet18 model for 100 epochs with batch size of 4.
Then we split the unseen datasets into $\mathbb{D}_{\text {train }}$, $\mathbb{D}_{\text {test}}$ and $\mathbb{D}_{\text {val}}$. For each category of the flower dataset, the number of train, test, and valid images are 10, 15, and 15, respectively. For each category of Animal Faces and VGGFace dataset, the number of train, test and valid images are 30, 35, and 35, respectively.
We train a new classifier using the pre-trained model on seen classes with $\mathbb{D}_{\text {train }}$ without any augmentation, which is denoted as ``Base''. Then we generate images to augment $\mathbb{D}_{\text {train }}$ with LoFGAN and our WaveGAN. The number of augmented images is 30 for Flower dataset and 50 for Animal Face and VGGFace datasets.
\begin{table}
\centering
\caption{\textbf{Classification results of augmentation.}}
\resizebox{.8\textwidth}{!}{
\begin{tabular}{c|c|c|c|c}
\toprule
Datasets                  & Base            & LoFGAN               & WaveGAN-M (ours)   & WaveGAN-B (ours)         \\ \hline
Flower                    & 64.71           & {\ul80.78}           & 70.20              & \textbf{84.71}           \\
Animals                   & 20.00           & 26.10                & {\ul31.81}         & \textbf{32.19}           \\
VGGFace                   & 50.76           & {\ul64.74}           & 62.96              & \textbf{77.36}           \\  \bottomrule
\end{tabular}
}
\label{tab:clsresults}
\end{table}

The classification results are given in Tab.~\ref{tab:clsresults}.
Compared with the results without any augmentation, our model achieves significant improvements, and our WaveGAN-B outperforms LoFGAN and WaveGAN-M obviously.
The effectiveness of using the generated images to augment the training dataset substantiates that our model can produce high-quality images, and the improvement on the classification accuracy provides a new data augmentation strategy for solving few-shot image classification problems.
\begin{figure*}
	\centering
	\includegraphics[width=\linewidth]{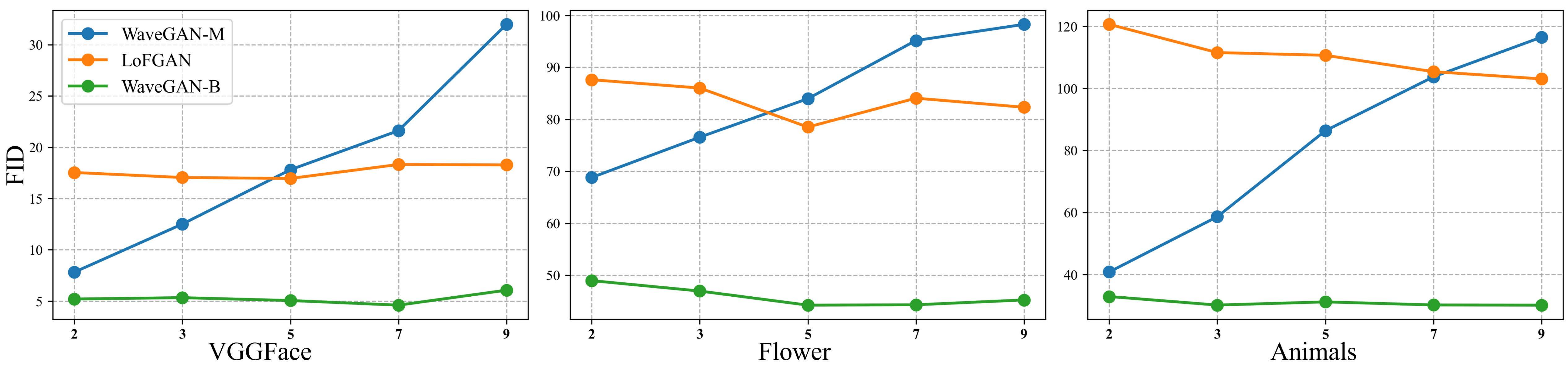}
	\caption{\textbf{Comparison results of our model under different shots of generation tasks.} The ordinate denotes FID scores, and the abscissa denotes different shots $K$.}
	\label{fig:IF_Shots}
\end{figure*}


\subsection{Influence of the Number of Shots}
All the experiments conducted before are 3-shot image generation tasks. We wonder to know the influence of different shots on our model.
We perform different shots of experiments with $K \in \{2, 3, 5, 7, 9\}$.
The number of training and testing images for different shots of the experiment are the same.
Fig.~\ref{fig:IF_Shots} demonstrates the performance of our WaveGAN and LoFGAN.
We can observe from the figure that when $K$ is relatively small, WaveGAN-M is better than LoFGAN. However, the performance of WaveGAN-M degrades with the number of images increases, making WaveGAN-M inadequate for image generation tasks with larger $K$.
Such phenomena corroborates our analysis in Sec.~\ref{sec:method} that the averaged transformation may fail to generalize to one specific image.
By contrast, our WaveGAN-B has relatively low sensitivity to $K$ and is shots-agnostic, thus WaveGAN-B has stronger generalization ability for different shots generation tasks, which manifests the superiority of our WaveGAN-B.


\section{Conclusion}
\label{sec:conclusion}
In this paper, we propose WaveGAN, the first few-shot image generation model that ameliorates the synthesis quality from the frequency domain perspective.
The key ingredients of our method are a WaveEncoder and a WaveDecoder.
Our WaveEncoder performs wavelet transformation on the different levels of features to obtain frequency signals.
We feed the decomposed low-frequency signals to the behind layers of the encoder, and we feed the high-frequency signals to our WaveDecoder.
Our design mitigates the generator's struggles of generating rich details for images, especially when limited data are available.
We further perform frequency $L_1$-loss to maintain the frequency information of real images, facilitating the fidelity of generated images.
Experimentally, our WaveGAN yields significant improvements on three challenging datasets, and the visualization and downstream classification results demonstrate that our WaveGAN can produce realistic images.
Besides, the ablation studies suggest the efficacy of each component of our method and substantiate that our approach complements the existing local fusion-based strategy.
Hopefully, our WaveGAN may inspire researchers to explore few-shot image generation from the frequency domain perspective.

\textbf{Acknowledgments.}
This work is supported by Shanghai Science and Technology Program ``Distributed and generative few-shot algorithm and theory research'' under Grant No. 20511100600 and ``Federated based cross-domain and cross-task incremental learning'' under Grant No. 21511100800, Natural Science Foundation of China under Grant No. 62076094, Chinese Defense Program of Science and Technology under Grant No.2021-JCJQ-JJ-0041, China Aerospace Science and Technology Corporation Industry-University-Research Cooperation Foundation of the Eighth Research Institute under Grant No.SAST2021-007.

\bibliographystyle{splncs04}
\bibliography{sections/egbib}

\clearpage
\section*{Appendix}
\label{appendix}
This appendix provides the supplementary information that is not elaborated in the main paper:
Sec.~\ref{sec:implementApp} provides the implementation details of our model.
Sec.~\ref{sec:AblationApp} provides quantitative results of our ablation studies.
Sec.~\ref{sec:EfficactHFapp} shows the efficacy of different High-frequency components.
Finally, Sec.~\ref{sec:2ddwtapp} presents the 2D DWT Visualization results of the images generated by our WaveGAN.

\appendix
\section{Implementation Details.}
\label{sec:implementApp}
Our encoder consists of five convolutional blocks and four wavelet transformation blocks.
The five convolutional blocks contain one convolution layer, followed by batch normalization and Leaky-Relu activation.
Our decoder is symmetrical with four upsampling blocks and one output convolutional layer.
Each upsampling block includes upsample operation followed by one convolutional block.
We perform our wavelet transformation after each convolution block in the encoder and employ inverse transformation after each convolution block in the decoder.
Our discriminator is the same as LoFGAN~\cite{gu2021lofgan} with four residual blocks and two fully connected layers.

Adam optimizer~\cite{kingma2014adam} is used and we train our model for 100,000 iterations.
At the beginning of 50,000 iterations, the learning rates for both the generator and the discriminator are set to 1e-4, after 5000 iterations, the learning rates decay linearly to 0.
We set $\lambda _{{\rm{cls }}}^G = \lambda _{{\rm{cls }}}^D = \lambda _{Fre} = 1 $, we save the final checkpoint to synthesis images for evaluation.
The batchsize is set to 8, and we sample hundreds of $K$-shot image generation tasks from $\mathbb{C}_s$.
Our model is implemented in PyTorch framework and trained on 1 $\times$ NVIDIA GeForce RTX 3090 GPU.

\section{Quantitative Results of Ablation Studies}
\label{sec:AblationApp}
Here we provide the quantitative results of our ablation studies in Sec.~\ref{sec:ablation} of the main paper,  demonstrating the effectiveness of each component of the proposed WaveGAN.
We remove each component and keep other settings unchanged to validate the contributions of each components of our WaveGAN, namely 1) the low-frequency skip connection, 2) the high-frequency skip connection, and 3) frequency $L_1$-loss.
Besides, we remove the LoF module to investigate the influence of local fusion on our model.
The quantitative results are given in Tab.~\ref{tab:ablation}, from which we can observe that each component boosts the synthesis performance.
Combining with the visualization results in Fig.~\ref{fig:ablation} further reflects the effectiveness of our proposed method.

\begin{table}[]
\centering
\small
\caption{\textbf{Ablation studies of our WaveGAN.} We test the efficacy of each components for our two transformation techniques, \emph{i.e.}, WaveGAN-M and WaveGAN-B.}
\resizebox{\textwidth}{!}{
\begin{tabular}{clcccccc}
\toprule
\multirow{2}{*}{Conditions} & \multirow{2}{*}{Type} & \multicolumn{2}{c}{Flowers}                            & \multicolumn{2}{c}{Animal Faces}                        & \multicolumn{2}{c}{VGGFace} \\
&      & FID ({$\downarrow$})   & LPIPS ({$\uparrow$})       & FID ({$\downarrow$})   & LPIPS ({$\uparrow$})    & FID ({$\downarrow$})      & LPIPS ({$\uparrow$})               \\ \midrule
WaveGAN-M w/o LoF         & Fusion                & 82.18           & 0.3720          & 52.64          & 0.5071          & 10.96           & \textbf{0.3822}                     \\
WaveGAN-M w/o LL          & Fusion                & 72.35           & 0.3709          & 67.53          & 0.5061          & 11.34           & 0.3795                              \\
WaveGAN-M w/o HL          & Fusion                & 87.99           & 0.3783          & 105.47         & 0.4981          & 21.48           & 0.3017                              \\
WaveGAN-M w/o $L_1$ Loss  & Fusion                & 73.86           & 0.3767          & 62.13          & 0.5017          & 12.29           & 0.3041                              \\
\textbf{WaveGAN-M (Ours)} & Fusion                & 63.79           & 0.3709          & 50.98          & 0.5014          & 8.62            & \textbf{0.3822}                     \\ \midrule
WaveGAN-B w/o LoF         & Fusion                & 47.37           & 0.3733          & 32.35          & 0.5080          & 5.35            & 0.3308                              \\
WaveGAN-B w/o LL          & Fusion                & 44.31           & 0.3803          & 31.12          & 0.5013          & 5.55            & 0.3223                              \\
WaveGAN-B w/o HL          & Fusion                & 85.25           & 0.3788          & 108.82         & 0.5011          & 20.61           & 0.3021                              \\
WaveGAN-B w/o $L_1$ Loss  & Fusion                & 45.52           & 0.3792          & 31.45          & 0.5047          & 5.99            & 0.3253                              \\
\textbf{WaveGAN-B (Ours)} & Fusion                & \textbf{42.17}  & \textbf{0.3868} & \textbf{30.35} & \textbf{0.5076} & \textbf{4.96}   & 0.3255                              \\ \bottomrule
\end{tabular}
}
\label{tab:ablation}
\end{table}

\section{Efficacy of High-frequency Components}
\label{sec:EfficactHFapp}
We feed the combination of high-frequency components (\ie, LH, HL, HH) to the decoder in the main version of our WaveGAN.
Here we test their efficacy by feeding only one of each to the decoder on Flowers dataset, the results are given in Tab.~\ref{tab:ablationFre}.
The results indicate that each high-frequency component has almost the same contribution to the generation quality, and fusing all of them (Full) yields the best results.
\begin{table}[tb!]
\centering
\tiny
\caption{\textbf{Efficacy of LH, HL, and HH tested on Flowers.}}
\resizebox{.8\textwidth}{!}{
\begin{tabular}{cccccc}
\toprule
Method            &Metric               & Only LF       & Only HF               & Only HH          & Full   \\ \Xhline{0.2pt}
WaveGAN-M         &FID                  & 69.73         & 69.49                 & 69.60            & 63.79       \\
WaveGAN-M         &LPIPS                & 0.3706        & 0.3654                & 0.3714           & 0.3709       \\ \Xhline{0.2pt}
WaveGAN-B         &FID                  & 48.65         & 43.76                 & 42.75            & 42.17       \\
WaveGAN-B         &LPIPS                & 0.3774        & 0.3779                & 0.3863           & 0.3868       \\ \Xhline{0.4pt}
\end{tabular}
}
\label{tab:ablationFre}
\end{table}

\section{2D DWT Visualization of High-frequency Components}
\label{sec:2ddwtapp}
\begin{figure}
	\centering
	\includegraphics[width=.8\linewidth]{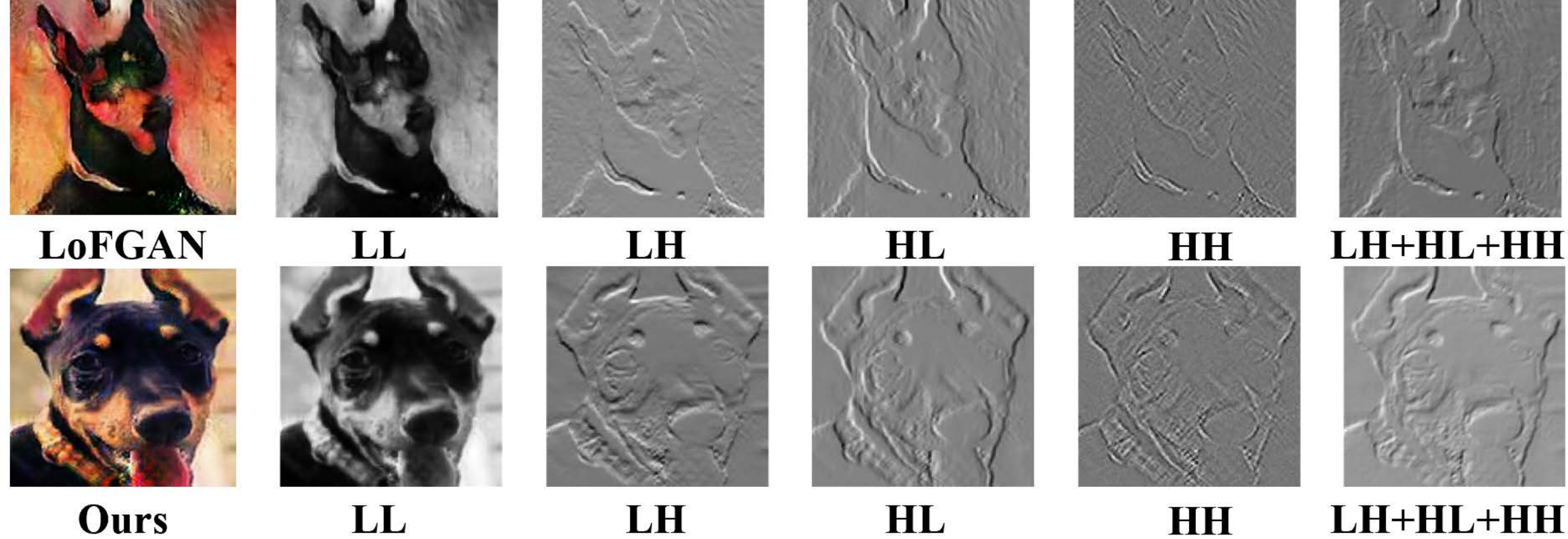}
	\caption{\textbf{2D DWT visualization results of frequency components.}}
	\label{fig:Wavelet2D}
\end{figure}

We provide the 2D DWT~\cite{liu2019attribute}~\cite{liu2018multi} visualization results of generated images in Fig.~\ref{fig:Wavelet2D}, which further demonstrates that our method is frequency-aware and produces images with informative frequency signals.

\end{document}